\documentclass[10pt,twocolumn,letterpaper]{article}

\usepackage{wacv}              

\usepackage{graphicx}
\usepackage{amsmath}
\usepackage{amssymb}
\usepackage{booktabs}
\usepackage[normalem]{ulem}
\usepackage[dvipsnames]{xcolor}
\usepackage{longtable}
\usepackage{makecell}
\usepackage{threeparttable}

\usepackage[pagebackref,breaklinks,colorlinks]{hyperref}

\usepackage[capitalize]{cleveref}
\crefname{section}{Sec.}{Secs.}
\Crefname{section}{Section}{Sections}
\Crefname{table}{Table}{Tables}
\crefname{table}{Tab.}{Tabs.}

\def\wacvPaperID{7} 
\def\confName{LLVM-AD}
\def\confYear{2025}

\newcommand{\zd}[1]{\textcolor{brown}{[ZD: #1]}}
\newcommand{\yz}[1]{\textcolor{magenta}{[YZ: #1]}}
\newcommand{\yw}[1]{\textcolor{red}{[YW: #1]}}
\newcommand{\red}[1]{\textcolor{red}{#1}}
\renewcommand{\zd}[1]{}
\renewcommand{\yz}[1]{}
\renewcommand{\yw}[1]{}
\renewcommand{\red}[1]{}

\begin{document}
\title{FROST-Drive: Scalable and Efficient End-to-End Driving with a Frozen Vision Encoder}

\author{Zeyu Dong\\
Stony Brook University\\
{\tt\small Zeyu.Dong@Stonybrook.edu}
\and
Yimin Zhu\\
Stony Brook University\\
{\tt\small yimzhu@cs.stonybrook.edu}
\and
Yu Wu\\
Rutgers University\\
{\tt\small yu.wu@rutgers.edu}
\and
Yu Sun\\
Sunrise Technology Inc.\\
{\tt\small sunrisetechnology001@gmail.com}
}

\maketitle

\begin{abstract}
End-to-end (E2E) models in autonomous driving aim to directly map sensor inputs to control commands, but their ability to generalize to novel and complex scenarios remains a key challenge. The common practice of fully fine-tuning the vision encoder on driving datasets potentially limits its generalization by causing the model to specialize too heavily in the training data. This work challenges the necessity of this training paradigm. We propose FROST-Drive, a novel E2E architecture designed to preserve and leverage the powerful generalization capabilities of a pretrained vision encoder from a Vision-Language Model (VLM). By keeping the encoder's weights frozen, our approach directly transfers the rich, generalized world knowledge from the VLM to the driving task. Our model architecture combines this frozen encoder with a transformer-based adapter for multimodal fusion and a GRU-based decoder for smooth waypoint generation. Furthermore, we introduce a custom loss function designed to directly optimize for Rater Feedback Score (RFS), a metric that prioritizes robust trajectory planning. We conduct extensive experiments on Waymo Open E2E Dataset, a large-scale datasets deliberately curated to capture the long-tail scenarios, demonstrating that our frozen-encoder approach significantly outperforms models that employ full fine-tuning. Our results provide substantial evidence that preserving the broad knowledge of a capable VLM is a more effective strategy for achieving robust, generalizable driving performance than intensive domain-specific adaptation. This offers a new pathway for developing vision-based models that can better handle the complexities of real-world application domains.
\end{abstract}

\section{Introduction}
End-to-end approaches to autonomous driving have emerged as a promising research direction, aiming to directly map raw sensor inputs to vehicle control commands. This paradigm seeks to overcome the limitations of traditional modular pipelines, which can suffer from compounding errors and the need for extensive manual tuning of individual components \cite{bojarskiEnd2016,chenEndtoend2023}. The dominant paradigm in this domain has converged on a modularized E2E design, which typically consists of a powerful vision encoder, a planner, and a decoder that generates trajectory waypoints \cite{shaoSafetyEnhanced2022,shaoReasonNet2023,huPlanningoriented2023}. In most contemporary studies, this entire architecture, including the vision encoder, is trained jointly from end to end.

\begin{figure}[htb]
\centering
\includegraphics[width=\linewidth]{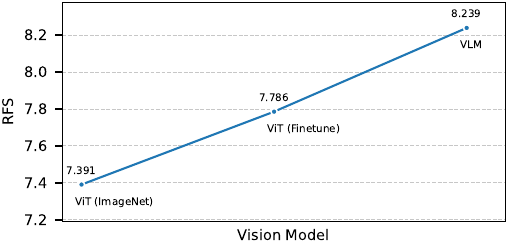}
\caption{Comparison of the model performance for different vision encoder approaches on Waymo E2E Dataset; ViT (ImageNet): use a frozen ViT pre-trained with ImageNet dataset; ViT (Finetune): fine-tune the E2E model end-to-end with Waymo Dataset; VLM: use a frozen ViT from a 72B VLM.}\label{fig:rfs_result}
\end{figure}

\begin{figure}[htb]
\centering
\includegraphics[width=\linewidth]{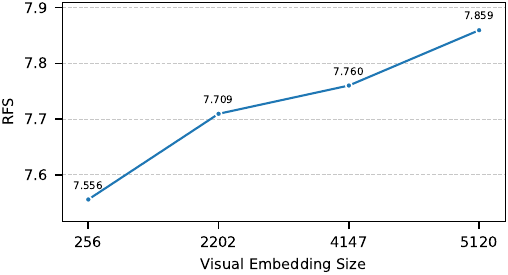}
\caption{Comparison of the model performance using different sizes of visual embeddings on Waymo E2E Dataset; X axis: end-to-end model using a ViT with different embedding sizes.}\label{fig:embedding_size}
\end{figure}

Fine-tuning or retraining the vision encoder has long been considered necessary for two main reasons. 1) There is a significant domain gap because models pre-trained on general datasets are not optimized for autonomous driving. Fine-tuning is thus required to adapt the feature extractor to recognize these domain-specific features; as shown in Figure~\ref{fig:rfs_result}, a fine-tuned Vision Transformer (ViT) improves model performance compared to a ViT pre-trained only on ImageNet. 2) The representational capacity of standard pre-trained models is often considered insufficient for the complexity of the driving task. As illustrated in Figure~\ref{fig:embedding_size}, performance scales with the embedding size, as larger embeddings carry richer information. Consequently, to utilize more powerful models with larger embeddings, retraining has been a prepared approach.

However, the practice of fully fine-tuning the vision encoder presents a fundamental trade-off that can harm a model's ability to generalize. While the process is computationally intensive, a more critical issue is that training end-to-end on a specific driving dataset limits the model's exposure to the long-tailed distribution of real-world events~\cite{hossain2024impact,jiaThink2023}. By intensely specializing in the most common scenarios within the training data, the model's broader, pre-trained knowledge can be compromised, weakening its ability to handle rare but critical situations. This creates a model that lacks the robust, generalizable understanding required for safe real-world deployment. This dilemma is a recurring challenge when adapting large-scale vision models to new application domains.


To resolve this, we propose a new approach that leverages the rich world knowledge embedded within a large-scale foundation model. Instead of training a vision encoder from scratch or extensively fine-tuning it, we adopt a powerful, pre-trained vision encoder and keep its weights frozen. The rationale is that vision encoders co-trained with language on vast datasets develop a richer, more contextual understanding of the world.
Additionally, these encoders are designed to generate high-dimensional feature embeddings capable of carrying the large volume of information required for complex driving scenarios. As a result, they are ideal for downstream tasks that require intricate reasoning. We argue that these two advantages—a better understanding of the world and a high-capacity feature representation—make it unnecessary to fine-tune the vision model.

We validate our hypothesis through extensive experiments on the Waymo Open E2E Dataset, a dataset corated to address long-tail scenarios in different environments. Our results demonstrate that a model utilizing a frozen vision encoder from a large foundation model significantly outperforms conventional approaches. Specifically, it surpasses the performance of both a fully fine-tuned model and models that use a frozen encoder pre-trained only on a general-purpose dataset. These findings suggest that retraining the vision component can be unnecessary for achieving state-of-the-art performance, provided that we transfers knowledge from a sufficiently powerful base model.

The primary contributions of this work are as follows:
\begin{enumerate}
\item We designed the FROST-Drive architecture for autonomous driving, featuring a frozen vision encoder, a transformer adapter, and a GRU decoder, all optimized with a custom RFS loss function. This approach serves as a generalizable and efficient alternative to full end-to-end training, achieving a top-three rank in the Waymo End-to-End Driving Challenge.
\item We provide critical insight into a common paradigm in end-to-end autonomous driving, showing that retraining a vision model may be unnecessary when the vision encoder is sourced from a sufficiently powerful VLM.
\item We are the first to conduct a comprehensive set of experiments analyzing how vision encoder size and feature embedding dimensionality affect the performance of a frozen-encoder approach, revealing a clear scaling relationship that validates the importance of using high-dimensional representations for this complex task.
\end{enumerate}

\section{Related Work}\label{sec:relatedwork}
\paragraph{Model Design in E2E Autonomous Driving} Unlike traditional modular approaches, end-to-end autonomous driving aims to generate driving actions directly from raw sensor data~\cite{chenEndtoend2023}. The field has flourished since NVIDIA introduced PilotNet~\cite{bojarskiEnd2016}, which utilized a simple 5-layer CNN as its image encoder. To handle more challenging driving scenarios, a variety of more complex architectures have been proposed. These include using ViT~\cite{sonataEndtoEnd2023} or ResNet~\cite{khanumEndtoEnd2020} as the image backbone, using swin transformer to generate video token for further text generation and control signal prediction~\cite{jin2023adaptactionawaredrivingcaption}, employing transformers for multi-sensor fusion~\cite{chittaTransFuser2022,shaoSafetyEnhanced2022}, incorporating the intermediate outputs and a holistic token mixer sub-network for effective feature adaptation~\cite{ding2024hintadholisticallyalignedinterpretability}, adopting Bird's-Eye View (BEV) image encoders~\cite{shaoReasonNet2023,chittaNEAT2021,huPlanningoriented2023}, and fine-tuning VLMs~\cite{shaoLMDrive2023}. Over time, the architectural design for end-to-end systems has largely converged on a modularized pipeline, typically featuring a vision backbone for feature extraction, a planning module, and a waypoint decoder~\cite{chenLearning2022,huPlanningoriented2023}. However, a common thread among most of these approaches is the requirement to fully fine-tune the entire model, including the large and computationally intensive vision module. This process demands extensive resources and large datasets, and it introduces a significant risk of overfitting to the training data.
\paragraph{Use of VLM in Autonomous Driving} The integration of VLMs has recently become a vibrant research frontier, leveraging their advanced reasoning and semantic understanding to enhance autonomous systems. Models like DriveGPT4 \cite{xuDriveGPT42023}, RAG-Driver \cite{yuanRAGDriver2024}, and DriveCoT \cite{wangDriveCoT2024} to generate natural language CoT reasoning for driving actions. Beyond reasoning in natural language, researchers are increasingly using VLMs for motion planning and control.
Some approaches employ hierarchical agents to enable real-world applications:
LM-Nav \cite{shahLMNav2022} and ViNG \cite{shahViNG2020} use a VLM as a visual guide for navigation. There are also dual-system designs where a high-level VLM makes strategic decisions that guide a low-level controller \cite{zhangADH2024,tianDriveVLM2024,dongGeneralizing2024}.
Others aim for a fully unified model;
for instance, GPT-Driver \cite{maoGPTDriver2023a} and EMMA \cite{hwangEMMA2024} represent a significant push towards all-in-one models that can process multimodal inputs and directly output driving commands.
A common theme across these E2E approaches, however, is the reliance on extensive fine-tuning to adapt the general-purpose VLM to the specific demands of E2E vehicle control.

\section{Methodolgy}\label{sec:methodology}
\begin{figure*}[t]
\centering
\includegraphics[width=0.9\textwidth]{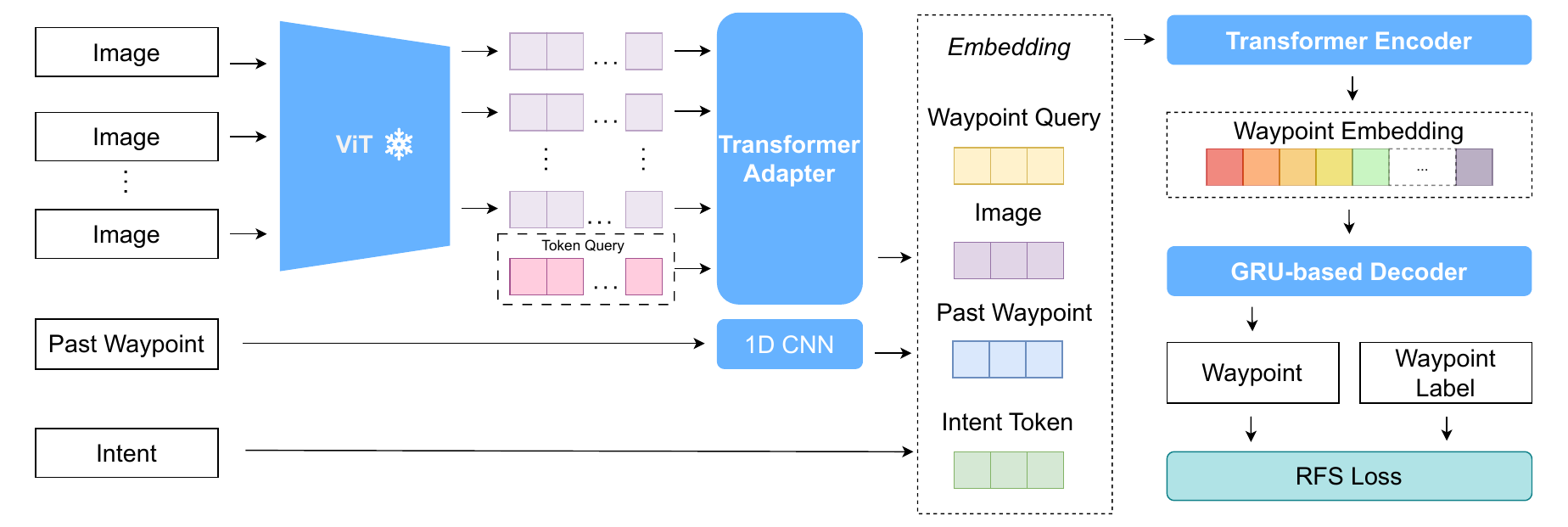}
\caption{Model Architecture}\label{fig:arch}
\end{figure*}
The primary objective of this work is to develop an end-to-end autonomous driving system that can safely and robustly navigate complex real-world environments. We formulate this as a trajectory planning task. Given a set of multi-view camera images $\mathcal{I} = \{I_{1}, \ldots , I_{N_c}\}$ from $N_c$ cameras at the current time $T$, a high-level, discrete driving intent $c$, a history of past states $S_{past} = (s_{T-k}, ..., s_{T-1})$ and current state $s_T$, where each \(s_i\) may contains several features like the position, speed, origination, etc. Our goal is to predict a sequence of local future waypoints $\hat{W} = (\hat{w}_{1}, ..., \hat{w}_{H})$ over a prediction horizon $H$ as a sequence of 2D-coordinates relative to the current location of the car. The core task is to learn a mapping function that generates a trajectory that is not only accurate but also safe and comfortable for passengers.

To achieve this, we propose a model designed for both high performance and training efficiency. As illustrated in Figure~\ref{fig:arch}, our architecture is composed of four primary components: a frozen, pretrained vision encoder from a VLM for feature extraction; a transformer-based adapter for fusing multi-camera input; a transformer encoder for fusing other multi-modal data; and a Gated Recurrent Unit (GRU) decoder for predicting the future waypoints. Another key aspect of our approach is the use of a novel loss function specifically designed to optimize for the \emph{Rater Feedback Score (RFS)}~\cite{waymo_e2e}, a critical metric for ensuring safe and reliable trajectory planning.

\subsection{Introduction to Rater Feedback Score}\label{sec:rfs}
In order to calculate the RFS, we define the reference trajectories, provided by dataset, as preferred rater specified trajectories, with a score $\bar{r}$ for each. Suppose the preferred rater specified trajectories are \(W = (w_{1}, w_{2}, ..., w_{H})\). And the predicted waypoints are \(\hat{W} = (\hat{w}_{1},\hat{w}_{2}, ..., \hat{w}_{H})\) as defined in previous paragraph.
With given preferred rater specified trajectories, we therefore define \emph{trust regions}. A trust region is defined for the region within a given lateral and longitudinal threshold of the rater-specified trajectory at a given time $t$ $(t= 3$s, $5$s are used in this work). The size of the trust region depends on the time step of the waypoint (\textbf{time-based thresholds}) and the velocity at the corresponding time (\textbf{speed-based scaling}) in both lateral and longitudinal directions and we finally derive the trust region by following steps:

\paragraph{Time-based thresholds.}  
The raw lateral/longitudinal thresholds $\tilde{\tau}_{\mathrm{lat}},\tilde{\tau}_{\mathrm{lng}}$ (in meters) are defined in Table \ref{tab: Time-based thresholds}.

\begin{table}[ht]
\centering
\begin{tabular}{ccc}
\toprule
Time $t$ & $\tilde{\tau}_{\mathrm{lat, t}}$ & $\tilde{\tau}_{\mathrm{lng, t}}$ \\
\midrule
3 & 1.0 & 4.0 \\
5 & 1.8 & 7.2 \\
\bottomrule
\end{tabular}
\caption{Trust region tolerance values over time}
\label{tab: Time-based thresholds}
\end{table}
\vspace{-2em}

\paragraph{Speed-based scaling.}  
The raw thresholds are scaled by the velocity(m/s) of the preferred rater specified trajectory:
{\small
\[
\mathrm{scale}(v)=
\begin{cases}
  0.5, & v<1.4,\\[6pt]
  0.5+0.5\times\frac{v-1.4}{11-1.4}, & 1.4\le v<11,\\[10pt]
  1, & v\ge 11.
\end{cases}
\]
}
The choice of speed for different scaling comes from the suggestions from~\cite{tefft2010car}. The speed 1.4m/s is around 5kmh (or 3 mph), in which pedestrian crash fatality is relatively low, while 11 m/s is around 40 kmh (or 25mph), in which pedestrian crash fatality reachs around 10\%. So we set up three intervals for the speed of safety, injury, fatality, accordingly.

\paragraph{Final thresholds.} Once we have the raw thresholds and the scale factors, the final thresholds are computed by
\(
\tau_{\mathrm{lat}}(t,v)=\mathrm{scale}(v)\,\tilde{\tau}_{\mathrm{lat,t}}
\)
and 
\(\tau_{\mathrm{lng}}(t,v)=\mathrm{scale}(v)\,\tilde{\tau}_{\mathrm{lng,t}}.\)

If the predicted waypoint is inside the trust region, RFS is assigned to be full score. However, if the predicted waypoint is outside of the trust region, we calculate an exponentially decreasing score for this waypoint. The overall formula is:
\begin{align*}
        \text{RFS}(w_t,\hat{w_t})=
        \begin{cases}
          \bar{r}&,\quad \text{when} \quad\Delta\le 1,\\
          \displaystyle
          \bar{r}\times 0.1^{
            \Delta-1}&, \quad \text{otherwise}.
        \end{cases}
\end{align*}
where $\Delta \doteq\max\left\{\dfrac{\Delta_{\mathrm{lat,t}}}{\tau_{\mathrm{lat,r}}},\dfrac{\Delta_{\mathrm{lng,t}}}{\tau_{\mathrm{lng,t}}}\right\}$ is the maximum distance error among lateral or longitudinal directions, $\Delta_{\mathrm{lat, t}}$ and $\Delta_{\mathrm{lng,t}}$ are lateral and longitudinal distance errors between the predicted waypoint \(\hat{w}_t\) and corresponding waypoint \(w_t\) in preferred rater specified trajecotory at time $t$, and $\bar{r}$ is the full score of preferred rater specified trajectory.

Besides RFS, Average Distance Error (ADE) is used secondary metric in this work. For each predicted trajectory, ADE is defined as the Euclidean distance between the prediction and the highest-scored rater trajectory at each time step and calculate average. Compared to conventional metrics like ADE, RFS is customized for more realistic autonomous driving scenarios. It is speed-aware, allowing for larger tolerable errors at higher speeds to reflect real-world physics. It also features deviation tolerance, defining a "trust region" that rewards any safe trajectory, not just perfect imitation of a single path. Finally, RFS has a strong emphasis on safety, as the score decays exponentially for any waypoint outside this region, heavily penalizing potentially dangerous deviations.

\subsection{Model Architecture}\label{sec:model_architecture}
As illustrated in Figure~\ref{fig:arch}, the architecture is composed of four primary components: a frozen, pretrained vision encoder from a VLM for feature extraction; a transformer-based adapter for fusing multi-camera input; a transformer encoder for fusing other multi-modal data; and a GRU decoder for predicting the future waypoints.

\paragraph{Pretrained Vision Encoder}
The foundation of our model is a powerful, frozen vision encoder, denoted as $f_{ViT}$, which is sourced from InternVL3 \cite{chen2024expanding, wang2024mpo, chen2024far, chen2024internvl}, a state-of-the-art VLM. InternVL3 excels in multimodal perception and reasoning, making its vision encoder uniquely suited for complex scene understanding. The vision model uses a pixel-unshuffle technique to support high-resolution image data while producing a high-dimensional embedding.

For our task, each camera view $I_{j} \in \mathcal{I}$ in the multi-camera setup is resized to 448x448 pixels. Each image $I_{j}$ is then processed independently by the encoder to extract a corresponding high-dimensional image feature, or visual embedding, $E_{img,j} = f_{ViT}(I_{j})$. The resulting embedding has a shape of $L_{img} \times d_{img}$, where $L_{img}$ is the number of visual tokens and $d_{img}$ is the embedding dimension. By keeping the encoder weights frozen, we leverage the rich world knowledge learned during its extensive pretraining, avoiding the computational cost and overfitting risks associated with fine-tuning.

\paragraph{Transformer Adapter}
Inspired by the BLIP-2 architecture \cite{liBLIP22023}, we employ a transformer-based adapter that uses a set of learnable query tokens to efficiently fuse information from the multi-view cameras and condense the visual information. First, the high-dimensional visual embedding $E_{img,j}$ from each camera view $I_j$ is passed through a linear projection layer, $f_{proj}$, to reduce its feature dimension from $d_{img}$ to a smaller dimension $d_{model}$. This step, producing $E'_{img,j} = f_{proj}(E_{img,j})$, significantly reduces the number of parameters in the trainable parts of our model.

Next, the projected visual tokens from all $N_c$ camera views are concatenated along the token dimension to form a single feature sequence: $E'_{img} = [E'_{img,1}; E'_{img,2}; \ldots; E'_{img,N_c}]$. To compress this long sequence, we utilize a set of $L_{\mathcal{I}}$ learnable query tokens, $Q_{\mathcal{I}} \in \mathbb{R}^{L_{\mathcal{I}} \times d_{model}}$. We perform cross-attention where $Q_{\mathcal{I}}$ acts as the query and $E'_{img}$ serves as both the key and value. This allows the model to distill the most salient information from the entire visual context into a compact representation
\[ E_{\mathcal{I}} = \text{CrossAttn}(Q_{\mathcal{I}}, E'_{img}, E'_{img}).\]
The resulting fused image embedding, $E_{\mathcal{I}}$, has a fixed shape of $L_{\mathcal{I}} \times d_{model}$, where $L_{\mathcal{I}} < N_c \times L_{img}$. This compression is crucial for computational efficiency in the downstream planning module.

\paragraph{Intent and Past State Embedding}
The driving decision depends not only on the camera images but also on the driving intent and the vehicle's past state. To fuse this information, we convert both the intent and past state into single-token embeddings of size $d_{model}$. For the categorical intent input, we apply one-hot encoding followed by a linear layer to get the intent embedding $E_{c}$. For the vehicle's past state data, we leverage its temporal dependency using a 1D Convolutional Neural Network (CNN). Specifically, the past states are concatenated along the temporal dimension, $S_{past} = [s_{T-k}; ...; s_T]$, and we apply a multi-layer 1D CNN followed by a max-pooling layer to get the past state embedding
\[E_{s} = \text{MaxPool}(\text{CNN}(S_{past})).\]
Both $E_{c}$ and $E_{s}$ are single-token embeddings of size $d_{model}$.

\paragraph{Transformer-based Planner}
To fuse the multimodal inputs of camera images, driving intent, and the vehicle's past state into a unified representation for planning, we use a transformer encoder architecture. The input to this encoder is a sequence formed by concatenating the fused camera embedding, the intent embedding, the past state embedding, and a learnable waypoint query token: $E_{in} = [E_{\mathcal{I}}; E_{c}; E_{s}; Q_{wp}]$. The transformer encoder processes this sequence through several layers of self-attention, allowing all modalities to interact and exchange information. The final output embedding corresponding to the waypoint query token, $E_{wp}$, serves as a context-rich feature vector that informs the vehicle's future trajectory.
\begin{align*}
& E_{out} = \text{SelfAttn}(E_{in}, E_{in}, E_{in}). \\
& E_{wp} = E_{out,l_{wp}},
\end{align*}
where \(l_{wp}\) is the corresponding token position for \(Q_{wp}\). The embedding \(E_{wp}\) encapsulates the comprehensive driving context required for the final decoding step.

\begin{figure}[ht]
\centering
\includegraphics[width=0.7\linewidth]{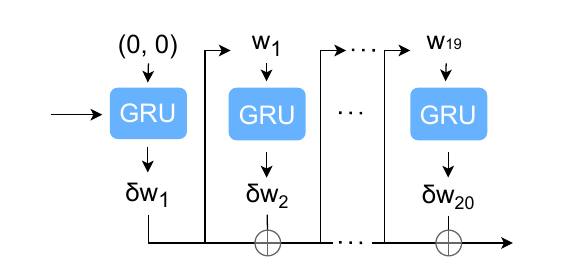}
\caption{GRU-based Decoder Architecture.}\label{fig:gru-decoder}
\end{figure}

\paragraph{GRU-based Waypoint Decoder}
Following established best practices \cite{chittaTransFuser2022}, we utilize a GRU-based decoder illustrated in Figure~\ref{fig:gru-decoder} to generate a smooth and temporally coherent local future trajectory. The waypoint feature vector produced by the planner serves as the initial hidden state for the GRU: $h_0 = E_{wp}$. We set the initial waypoint $w_0 = (0,0)$. The decoder then operates autoregressively, predicting the delta between the current and next waypoint at each step
\[(h_t, \delta w_t) = \text{GRU}(h_{t-1}, w_{t-1}),\]
and the next waypoint is calculated as $w_t = w_{t-1} + \delta w_t$. This flow-based prediction ensures smooth control outputs, which is crucial for the robustness of the waypoint prediction.

\subsection{Loss Function Design}\label{sec:loss_function}
To directly optimize for a robust and safe trajectory, we introduce a surrogate loss function based on the RFS metric defined in Section~\ref{sec:rfs}. Instead of the standard Mean Squared Error (MSE) which penalizes the $L_2$ distance, or the widely used $L_1$ loss~\cite{liEgo2024,zhaiRethinking2023}, our loss function focuses on minimizing the maximum weighted displacement error along the vehicle's lateral and longitudinal axes. For the $t$-th waypoint, the loss is defined as:
\[
\mathcal{L}_0(w_t, \hat{w}_t) = \max \left\{ \frac{\Delta_{\text{lat},t}}{\tau_{\text{lat},t}}, \frac{\Delta_{\text{lng},t}}{\tau_{\text{lng},t}} \right\}
\]
Here, $\Delta_{\text{lat},t}$ and $\Delta_{\text{lng},t}$ represent the lateral and longitudinal errors for waypoint $t$, while $\tau_{\text{lat},t}$ and $\tau_{\text{lng},t}$ are their corresponding tolerance thresholds as defined in Section~\ref{sec:rfs}. The final loss is then computed as the average over all waypoints in the horizon $H$:
\[\mathcal{L}(W, \hat{W}) = \dfrac{1}{H} \sum_{t=1}^{H} \mathcal{L}_0(w_t, \hat{w}_t).\]
This loss function can be interpreted as a combination of the $L_\infty$ and $L_1$ norms. For each 2D waypoint, we use the $L_\infty$ norm on the weighted axis-wise errors, which encourages robustness by penalizing the worst-case deviation. The summation over the sequence of waypoints then acts like an $L_1$ loss, which is known to be more robust to outliers~\cite{qifakeRobust2005}. It is generally considered that a pure $L_\infty$ objective can slow down convergence; our combined approach mitigates this while retaining the benefits of robustness.

Furthermore, before calculating the maximum displacement, we perform a coordinate transform to align with the vehicle's orientation and apply different weights to the lateral and longitudinal errors based on the vehicle's speed. This makes the loss function more suitable for realistic driving scenarios, leading to more reliable driving behavior.

\section{Experiment}
We perform experiments on the Waymo End-to-End Driving dataset to validate our approach. Through these experiments, we demonstrate that leveraging a frozen visual encoder from a VLM yields superior performance in the end-to-end driving task. We show that this performance advantage is attributed to the following two points:
\begin{enumerate}
\item The frozen vision encoder from the VLM inherits a rich world knowledge. This enables the model to extract more driving-relevant features.
\item The encoder's ability to process high-resolution inputs allows it to generate high-dimensional feature embeddings. These rich representations are essential for capturing the complexity of the driving task.
\end{enumerate}

\subsection{Dataset}\label{sec:dataset}

Our study is conducted using the Waymo Vision-based End-to-End Driving dataset, a large-scale datasets deliberately curated to capture these long-tail scenarios. It covers diverse environments and rare events occur with a frequency of less than 0.003\% in daily driving\cite{waymo2025e2edriving}, such as navigating construction during public gatherings, avoiding fallen pedestrians, and handling unexpected freeway obstacles. This makes the dataset a challenging and valuable benchmark for advancing generalizable autonomous end-to-end driving capabilities, especially for the task of end-to-end future waypoint planning that is a significantly harder problem than classical perception tasks.

\begin{figure}[ht]
    \centering
    \includegraphics[width=0.9\linewidth]{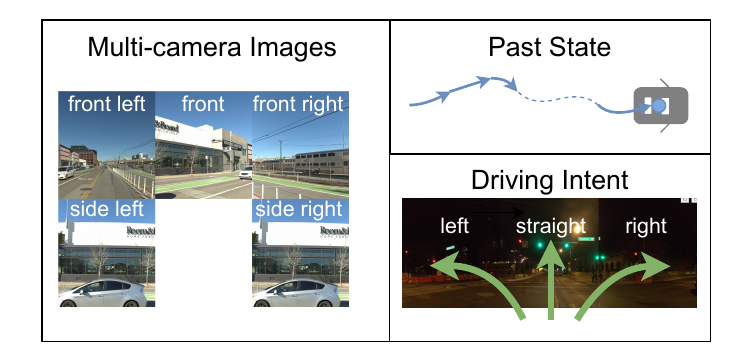}
    \caption{Example of raw inputs provided in the dataset.}
    \label{fig:raw-input}
\end{figure}

This dataset comprises 4021 unique driving scenes, with each scenes capturing a continuous 20-second-long scene sampled at 4Hz. The task is to predict the vehicle's future trajectory based on its current sensor inputs and past state. As illustrated in Figure~\ref{fig:raw-input}, the inputs provided are

\begin{itemize}
\item \textbf{Multi-camera Images ($\mathcal{I}$):} A set of high-resolution videos from 8 cameras providing a 360-degree field of view. We use a subset of \(N_c = 5\) cameras that provide a comprehensive forward and side-facing view: \textit{front, front-left, front-right, side-left, and side-right}.
\item \textbf{Past State ($S_{past}$):} The vehicle's location, velocity, and acceleration for the past 4 seconds, forming a sequence of length 16.
\item \textbf{Driving Intent ($c$):} A high-level command from the set $\{\texttt{go left},\ \texttt{go straight},\ \texttt{go right}\}$, which the vehicle should follow in a context-appropriate manner.
\end{itemize}
The model's objective is to output the future waypoints $\hat{W}$ for the next 5 seconds, corresponding to a prediction horizon of $H=20$.
 
\subsection{Evaluation Metrics}
The dataset is partitioned into training, validation, and testing sets. Our primary evaluation is conducted in an open-loop setting on the validation set. We use the RFS, as described in Section~\ref{sec:rfs}, to evaluate the quality of the predicted trajectory. To compute the RFS for the validation set, we use the ground-truth future waypoints as the reference trajectory, with a default score of $\bar{r} = 10$. Additionally, we report the ADE at 3 and 5 seconds as a supplementary metric.

\subsection{Experimental Setup}
Our experiments compare several vision backbones to isolate the effect of the pre-trained encoder. For all models tested, the vision encoder processes each camera frame to produce an embedding with a token length of $L_{img} = 256$. After the transformer adapter fuses the multi-camera views, the resulting token count is $L_{\mathcal{I}} = 256$. The key configurations of the VLM we use, detailing the relationship between their total parameters and the properties of their respective vision encoders, are summarized in Table~\ref{tab:model_configs}.

\begin{table}[htb]
\centering
\begin{tabular}{ccc}
\toprule
\textbf{VLM} & \textbf{Vision Encoder \#Params} & \boldmath{\(d_{img}\)} \\
\midrule
InternVL3-1B      & 300M      & 896       \\
InternVL3-14B     & 300M      & 5120      \\
InternVL3-38B     & 6B        & 5120      \\
InternVL3-78B     & 6B        & 8192      \\
\bottomrule
\end{tabular}
\caption{Configurations of the foundation models and their corresponding vision encoders used in our experiments.}\label{tab:model_configs}
\end{table}

\subsection{Results and Analysis}

\begin{figure*}[t]
\centering
\includegraphics[width=0.45\linewidth]{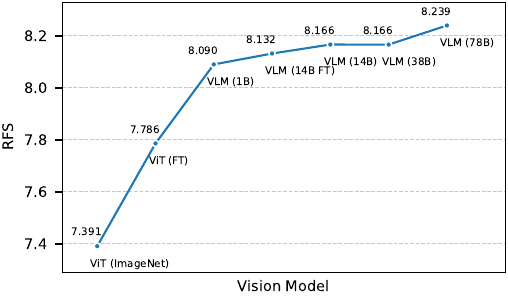}
\hspace{1em}
\includegraphics[width=0.45\linewidth]{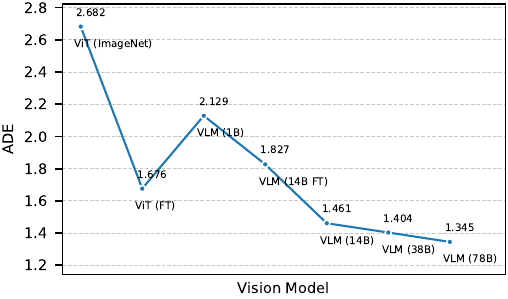}
\caption{Comparison of the model performance metric for RFS and ADE for different vision encoder approaches.}\label{fig:vlm_comparison_plots}
\end{figure*}

\paragraph{Impact of VLM Pre-training}
To validate that knowledge from a large pre-trained model provides a superior foundation for driving, a comparative analysis of several vision encoder configurations was conducted. Our proposed approach, using frozen encoders from foundation models of varying sizes (1B, 14B, 38B, and 78B parameters), was evaluated against three strong baselines: a frozen, ImageNet-pretrained ViT; a standard, fully fine-tuned ViT; and the 14B VLM encoder with its weights unfrozen for full fine-tuning. For a fair comparison, the baseline ViTs have 300M parameters, matching the vision encoder size of the 1B and 14B VLMs.

The results, summarized in Table~\ref{tab:vlm_comparison} and visualized in Figure~\ref{fig:vlm_comparison_plots}, provide compelling evidence for our hypothesis. First, all frozen VLM encoders significantly outperform both the ImageNet-pretrained and the fully fine-tuned ViT baselines, particularly on the RFS metric, which is more aligned with driving robustness. Second, performance consistently improves as the size of the source VLM scales from 1B to 78B. In addition, the fine-tuned 14B VLM performs worse than its frozen counterpart. This suggests that fine-tuning can degrade the rich, generalizable world knowledge learned during pre-training, forcing the model to over-specialize. This confirms that preserving this knowledge is more beneficial than intensive domain adaptation.

\begin{table}[htb]
\centering
\begin{threeparttable}[b]
\begin{tabular}{llll}
\toprule
\textbf{Models} &                     \textbf{RFS$\uparrow$} &                  \textbf{ADE@3s$\downarrow$} &                  \textbf{ADE@5s$\downarrow$} \\
\midrule
ViT (ImageNet) &  \makecell{7.39 \\ ($\pm$0.073)} &   \makecell{2.28 \\ ($\pm$0.05)} &  \makecell{3.08 \\ ($\pm$0.075)} \\
ViT (FT\tnote{1} ) &  \makecell{7.79 \\ ($\pm$0.015)} &   \makecell{1.2 \\ ($\pm$0.004)} &  \makecell{2.15 \\ ($\pm$0.009)} \\
VLM (14B FT\tnote{1} ) & \makecell{8.13 \\ ($\pm$0.014)} & \makecell{1.47 \\ ($\pm$0.004)} & \makecell{2.19 \\ ($\pm$0.009)} \\
VLM (1B)       &  \makecell{8.09 \\ ($\pm$0.015)} &  \makecell{1.84 \\ ($\pm$0.007)} &  \makecell{2.42 \\ ($\pm$0.011)} \\
VLM (14B)      &  \makecell{8.17 \\ ($\pm$0.015)} &  \makecell{1.04 \\ ($\pm$0.004)} &  \makecell{1.88 \\ ($\pm$0.009)} \\
VLM (38B)      &  \makecell{8.17 \\ ($\pm$0.015)} &  \makecell{0.98 \\ ($\pm$0.004)} &   \makecell{1.83 \\ ($\pm$0.01)} \\
VLM (78B)      &  \makecell{8.24 \\ ($\pm$0.014)} &  \makecell{0.95 \\ ($\pm$0.004)} &  \makecell{1.74 \\ ($\pm$0.009)} \\
\bottomrule
\end{tabular}
\begin{tablenotes}
\item[1]{FT: Finetuning}
\end{tablenotes}
\caption{Result compare E2E model with different vision encoders.}\label{tab:vlm_comparison}
\end{threeparttable}
\end{table}

\paragraph{Impact of Embedding Dimensionality}

To show that high-dimensional features are critical for this task, we performed experiment on the feature embedding size. For this experiment, we used the high-performing frozen encoder from the 38B VLM and progressively reduced the dimensionality of its output visual embedding before passing it to the downstream planner.

As detailed in Table~\ref{tab:embedding_ablation} and Figure~\ref{fig:embedding_size}, the results show a strong correlation between embedding size and driving performance. As the feature dimension is reduced from 5120 down to 256, both the RFS and ADE metrics degrade substantially. This indicates that both the quality of the model's learned knowledge and the richness of its feature representation are key drivers of performance in autonomous driving.

\begin{table}[htb]
\centering
\begin{tabular}{llll}
\toprule
\textbf{Embedding Size} &                     \textbf{RFS$\uparrow$} &                  \textbf{ADE@3s$\downarrow$} &                  \textbf{ADE@5s$\downarrow$} \\
\midrule
256  &  \makecell{7.68 \\ ($\pm$0.018)} &  \makecell{0.96 \\ ($\pm$0.005)} &  \makecell{2.02 \\ ($\pm$0.012)} \\
2202 &  \makecell{7.57 \\ ($\pm$0.018)} &  \makecell{1.16 \\ ($\pm$0.006)} &  \makecell{2.32 \\ ($\pm$0.014)} \\
4147 &  \makecell{7.81 \\ ($\pm$0.018)} &  \makecell{0.98 \\ ($\pm$0.005)} &  \makecell{1.99 \\ ($\pm$0.012)} \\
5120 &  \makecell{8.17 \\ ($\pm$0.015)} &  \makecell{0.98 \\ ($\pm$0.004)} &   \makecell{1.83 \\ ($\pm$0.01)} \\
\bottomrule
\end{tabular}
\caption{Comparison of the model performance when the visual embedding size changes.}\label{tab:embedding_ablation}
\end{table}


\begin{table*}[t]
\centering
\resizebox{\textwidth}{!}{%
\begin{tabular}{lrrrrrrrrrrr}
\toprule
{} & \multicolumn{11}{c}{\textbf{Categorical RFS}} \\
\cmidrule{2-12}
\textbf{Method Name} & \textbf{Spotlight}$\uparrow$ & \textbf{Constr} & \textbf{Inter} & \textbf{Ped} & \textbf{Cyclist} & \textbf{Multi} & \textbf{Single} & \textbf{Cut-ins} & \textbf{FOD} & \textbf{Special} & \textbf{Others} \\
\midrule
RAP & 8.6939 & 8.1798 & 8.0336 & 7.7604 & 8.0999 & 8.5232 & 8.0976 & 8.2741 & 7.8176 & 7.2041 & 7.7893 \\
\textbf{Ours} & 7.0941 & 8.2510 & 7.9658 & 7.9293 & 7.7192 & 7.6509 & 8.3287 & 7.9323 & 8.1297 & 8.0375 & 7.3775 \\ 
UniPlan & 6.9174 & 8.5600 & 7.8639 & 7.6384 & 7.7559 & 7.6699 & 8.1599 & 7.7859 & 8.0847 & 7.6702 & 7.4685 \\
Poutine & 6.8929 & 8.3595 & 8.1356 & 7.9529 & 7.8325 & 7.7789 & 8.6101 & 8.2588 & 8.2043 & 8.2965 & 7.5235 \\
HMVLM & 6.7269 & 8.6663 & 7.9043 & 7.8578 & 7.3925 & 7.5607 & 8.3563 & 7.5826 & 7.8842 & 7.9710 & 7.2013 \\
ViT-Adapter-GRU & 6.6722 & 8.4630 & 8.0471 & 7.8904 & 7.8346 & 7.6132 & 8.3424 & 7.9308 & 8.0682 & 8.0920 & 7.3889 \\
\bottomrule
\end{tabular}
}
\caption{Detailed Categorical RFS on the Waymo E2E Driving Challenge Leaderboard (as of 12/16/25).}\label{tab:leaderboard_categorical}
\end{table*}

\subsection{Qualitative Analysis}
\begin{figure}[htb]
\centering
\includegraphics[trim={0 0 0 400px},clip,width=0.8\linewidth]{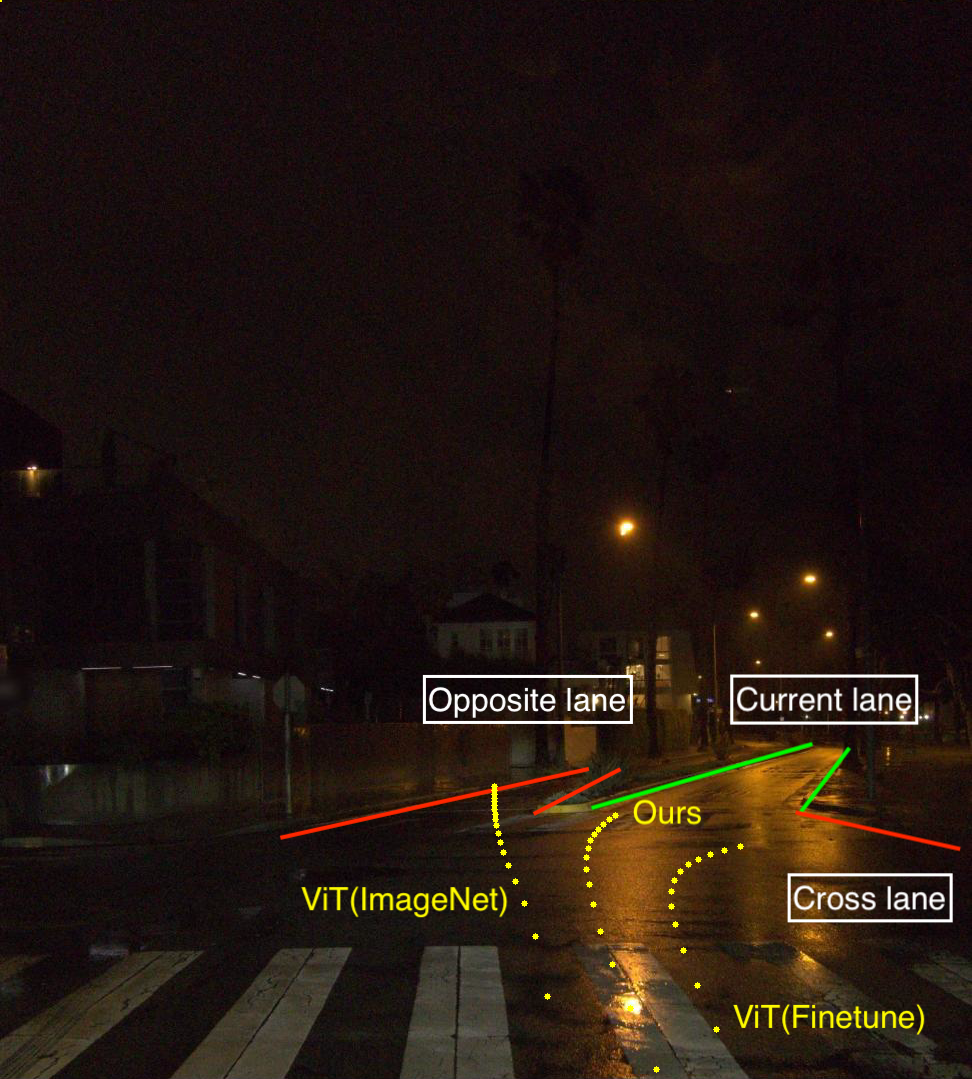}
\caption{Visualization of model performance in a complex intersection. Our model successfully identifies the correct driving path and navigates towards it, while the baseline model fails to interpret the scene correctly.}\label{fig:qualitative_comparison}
\end{figure}

To provide intuition for the quantitative improvements, we visualize a driving scenario in Figure~\ref{fig:qualitative_comparison}. This figure shows the front camera inputs and compares the predicted future trajectory from our FROST-Drive model against the baseline.

The selected scenario involves navigating a complex intersection where the correct driving path is the one shifted slightly to the right. Our FROST-Drive model demonstrates superior scene understanding; it not only identifies the visible lane but also correctly infers that the intended path is the shifted one. In contrast, the baseline models, lacking this deeper contextual awareness, fail to identify the correct path and follow the incorrect lane. This demonstrates a lack of true scene understanding.

This example highlights a key advantage of our approach: the frozen encoder's rich world knowledge allows it to better understand complex scenes beyond simple feature extraction. This leads to more robust, context-aware, and human-like driving behavior, particularly in non-standard situations where basic visual cues can be misleading.

\subsection{Result in Waymo Challenge Leaderboard}
To validate the effectiveness of our approach against the state of the art, we submitted our best-performing model to the official Waymo End-to-End Driving Challenge. As shown in Table~\ref{tab:leaderboard_main}, our method achieves a highly competitive third-place rank on the overall RFS metric. While the ADE is higher than some other methods, this is an expected trade-off resulting from our model's explicit optimization of the custom RFS loss function, which prioritizes robust, safe trajectories over exact path imitation.

As shown in Table~\ref{tab:leaderboard_categorical}, our model ranks second in the ``Spotlight'' category, a set of manually selected challenging edge-case scenarios~\cite{waymo2025e2edriving}. This confirms that our approach is powerful and efficient in generalization.


\begin{table}[htb]
\centering
\resizebox{\linewidth}{!}{%
\begin{threeparttable}[b]
\begin{tabular}{lrrr}
\toprule
\textbf{Method Name} & \textbf{RFS (Overall)$\uparrow$} & \textbf{ADE@5s$\downarrow$} & \textbf{ADE@3s$\downarrow$} \\
\midrule
RAP & 8.0430 & 2.6457 & 1.1741 \\
Poutine & 7.9860 & 2.7419 & 1.2055 \\
\textbf{Ours\tnote{1}} & 7.8560 & 3.5653 & 2.5373 \\
ViT-Adapter-GRU  & 7.8493 & 2.8888 & 1.4434 \\
UniPlan & 7.7795 & 2.8423 & 1.2671 \\
HMVLM & 7.7367 & 3.0715 & 1.3269 \\
\bottomrule
\end{tabular}
\begin{tablenotes}
\item[1]{Our result is hidden from the official leaderboard to comply with the double-blind policy.}
\end{tablenotes}
\end{threeparttable}
}
\caption{Overall Performance Comparison on the Waymo E2E Driving Challenge Leaderboard.}\label{tab:leaderboard_main}
\end{table}

\section{Conclusion}
In this work, we challenged the prevailing paradigm that state-of-the-art performance in end-to-end autonomous driving requires extensive fine-tuning of the vision encoder. We introduced FROST-Drive, a highly effective and efficient approach that leverages a frozen, pre-trained vision encoder from a VLM. Our method achieves superior performance while avoiding the high computational costs and overfitting risks associated with full model training.

Our experiments on the Waymo End-to-End Driving dataset provide compelling evidence for this approach. The results show that FROST-Drive surpasses a fully fine-tuned baseline but. In addition, its performance consistently improves with both the size of the VLM and the dimensionality of its feature embeddings. These findings validate our core hypothesis: the rich world knowledge and high-capacity features from large pre-trained models are critical for navigating complex driving scenarios. Our qualitative analysis further revealed that this enhanced understanding allows our model to handle challenging situations with greater robustness than baseline models. The effectiveness of our method was ultimately confirmed by its top-three ranking on the official Waymo Challenge leaderboard.

{\small
\bibliographystyle{ieee_fullname}
\bibliography{yimin,egbib,yuwu}
}

\end{document}